\documentclass{article}

\usepackage{adjustbox}
\usepackage{amsfonts}
\usepackage{amsmath}
\usepackage{bm}
\usepackage{bbm}
\usepackage{caption}
\usepackage{graphicx}
\usepackage[pagebackref]{hyperref}
\usepackage{cleveref}
\usepackage{multirow}
\usepackage{tikz}
\usepackage[utf8]{inputenc}
\usepackage[a4paper, total={6in, 8.5in}]{geometry}

\DeclareMathOperator*{\argmax}{argmax}

\usepackage{xcolor}

\title{DNA: Dynamic Network Augmentation}
\author{Scott Mahan$^1$ \hspace{1cm} Tim Doster$^2$ \hspace{1cm} Henry Kvinge$^2$ \\~\\ $^1$University of California, San Diego \hspace{1cm} $^2$ Pacific Northwest National Laboratory}
\date{}

\begin{document}

\maketitle

\begin{abstract}
    In many classification problems, we want a classifier that is robust to a range of non-semantic transformations. For example, a human can identify a dog in a picture regardless of the orientation and pose in which it appears. There is substantial evidence that this kind of invariance can significantly improve the accuracy and generalization of machine learning models. A common technique to teach a model geometric invariances is to augment training data with transformed inputs. However, which invariances are desired for a given classification task is not always known. Determining an effective data augmentation policy can require domain expertise or extensive data pre-processing. Recent efforts like AutoAugment optimize over a parameterized search space of data augmentation policies to automate the augmentation process. While AutoAugment and similar methods achieve state-of-the-art classification accuracy on several common datasets, they are limited to learning one data augmentation policy. Often times different classes or features call for different geometric invariances. We introduce Dynamic Network Augmentation (DNA), which learns input-conditional augmentation policies. Augmentation parameters in our model are outputs of a neural network and are implicitly learned as the network weights are updated. Our model allows for dynamic augmentation policies and performs well on data with geometric transformations conditional on input features. 
\end{abstract}

%\HK{DNA is a nice acronym, but would you want to consider a name that highlights the key feature of your method: that it is input conditional?}

\section{Introduction}

Deep neural networks are powerful classification models that often require large amounts of labeled data to be trained well. Data augmentation (DA) is the technique of augmenting a dataset with transformed inputs to artificially increase the quantity and diversity of training data \cite{KSH12}. Proper DA policies enhance the generalization of a machine learning model and can lead to more efficient training.

The transformations applied during DA can teach a model to be invariant to task-independent modes of variation within the data. For example, in image data, horizontal flips and other geometric transformations are often non-semantic for object classification. Many image classification tasks are insensitive to small color adjustments, and other data domains can be invariant to their own group actions (such as permutation of inputs). When training data is augmented using these transformations, the model can learn the corresponding invariances, resulting in better generalization at test time.

Some neural network architectures encode geometric invariances using built-in model constraints. Convolutional neural networks are approximately translation-invariant by design since the same filters are applied across various patches of an input image \cite{LB98}. Other architectures use weight-sharing and pooling operations to enforce invariance to rotations, permutations, or other group actions \cite{CW16a,CW16b,WC19}. Such models can be very efficient when these hard-coded invariances accurately reflect non-semantic transformations in the data. However, most of them are limited to full group invariances which may negatively impact performance (e.g., some labels in MNIST \cite{LCB10} are not invariant to 180-degree rotations). Some models allow for partial group actions \cite{BFIW20} by sampling from a subset of transformations (e.g., rotation between $-30$ and $30$ degrees), but even this DA framework can be too rigid for effective generalization. Moreover, hard-coding invariances requires domain knowledge of which transformations do not impact classification for a given dataset.

DA is an alternative to hard-coding invariance into a network architecture. It only requires that we can simulate the transformations with which we are trying to build invariance. Additionally, DA techniques are flexible in that they augment the training data according to some predetermined set of transformations that need not follow a specific algebraic structure. However, designing DA policies still requires a great deal of domain knowledge, as effective DA implementation can vary greatly from one data set to another. For example, image reflections can be an effective way to generate new data for CIFAR-10 \cite{K09}, but would likely reduce accuracy on MNIST where chirality plays an important role in digit identity. Thus, there is a need for new DA strategies which are tailored to a specific dataset.

Recent efforts have been made to optimize DA policies from training data, which would further automate the machine learning pipeline. These methods include generative DA \cite{LBC17,SPTSWW17,TPCPR17} as well as parameterized policy optimization \cite{CZMVL19,HLSAC19,LKKK19,LHWHRY20}. Automating the augmentation process eliminates the need for domain expertise while still allowing for great model performance. 

However, one limitation of all previously mentioned augmentation methods -- both manual and automated -- is that they learn one augmentation policy for an entire dataset. While this is a common approach in DA, the range and types of variation present differs across classes and even individual instances in a dataset. For example, if a dataset contains overhead images of airplanes, it is critical that a network learns rotational invariance to these since they have no preferred orientation. On the other hand, buildings do not often appear upside down, so learning rotation invariance for this class may be less important or may even harm performance. We address this problem with Dynamic Network Augmentation (DNA), which draws from the input-dependent nature of dynamic neural networks \cite{HHSYWW21} to optimize augmentation policies. Our model features an augmentation network that learns input-conditional parameterizations of DA policies. We then use a differentiable relaxation of augmentation sampling \cite{JGP17} before feeding the augmented input into a classification network. The result is a fully differentiable end-to-end model with learnable input-conditional data augmentation.

We test DNA on CIFAR-10, CIFAR-100 \cite{K09}, and SVHN \cite{NWCBWN11} %, and ImageNet [cite] 
by first training the augmentation network and then using the learned input-conditional DA policy to train the classification network. Our algorithm requires similar amounts of time for policy search and classifier training to the most recent automatic DA methods, and it achieves comparable accuracy to other methods on these datasets. Moreover, the input-conditional nature of our data augmentation allows us to explore invariances of our model on a finer scale.

\section{Related Work}

We focus on using DA for building a model that is robust to non-semantic geometric transformations, rather than using networks with hard-coded invariances. Manually selected DA policies for common datasets have existed for a long time, relying on domain expertise. For example, it has long been common practice to improve performance on MNIST by using geometric transformations like translation, scaling, shearing, and rotating \cite{SSP03}. On the other hand, natural image data like CIFAR-10 is typically augmented with cropping, mirroring, and color-changing operations to improve model generalization \cite{KSH12}. These are simple datasets for which DA has been studied extensively, but new datasets require their own expertise and investigation into effective augmentation strategies. Other augmentations that have found success in the image domain include Cutout \cite{DT17}, Mixup \cite{ZCDL18}, and CutMix \cite{YHOCCY19}. Each of these operations replaces or covers a random patch of the image in some way.

Approaches to automate DA generally fall into one of two categories. The first approach generates entirely new samples from existing training data. For example, Smart Augmentation \cite{LBC17} merges two samples from the same class to create new data. Other models use Generative Adversarial Networks (GANs) \cite{GPMXWOCB14} to generate augmented data by either refining synthetic images \cite{SPTSWW17} or sampling from the approximate training distribution using an expectation maximization algorithm \cite{TPCPR17}. Each of these methods seeks to generate new samples that are likely to come from the same distribution as the actual training examples, thus increasing the quantity and diversity of traning data.

The other approach to automated DA involves optimizing over a parameterized policy search space that includes a predetermined list of transformations. These algorithms are inspired by neural architecture search, where reinforcement learning or population based optimization is used to optimize model architectures from data \cite{SZ15,SLJSRAEVR15}. In the DA setting, optimization occurs over the parameterized policy search space, and this optimization is guided by evaluating DA policies on training data. AutoAugment (AA) \cite{CZMVL19} introduces a very general parameterized augmentation policy search space consisting of many common geometric and color transformations, allowing for robust DA that can be learned from training data. AA optimizes the augmentation policy using reinforcement learning, but proves very costly as each step of the reinforcement learning algorithm involves training a network on a newly augmented training set. Population Based Augmentation (PBA) \cite{HLSAC19} performs population based optimization on the DA policy, while Fast AutoAugment (Fast AA) \cite{LKKK19} uses Bayesian optimization. Both methods vastly improve the computation time of AA while obtaining similar accuracy on test data. To make DA policy search even more efficient, Differentiable Automatic Data Augmentation (DADA) \cite{LHWHRY20} relaxes policy selection to a differentiable process which can then be optimized using standard network backpropagation. DADA in particular draws from differentiable architecture search algorithms \cite{LSY19}. Each of these models performs well on image datasets without any prior knowledge of which augmentations are beneficial. 

Finally, our DNA model learns input-conditional DA policies with a separate augmentation network, allowing for dynamic data augmentation. We note that none of the aforementioned methods address the situation where the types of augmentation that should be applied depend on the particular instance at hand. This is the primary novelty of our proposed DNA method.

\section{Dynamic Network Augmentation (DNA)}

\begin{figure*}
	\centering
	\resizebox{\linewidth}{!}{
		\def\cnnsize{1.5}
		\def\cnnoo{2.0}
		\def\cnnot{-0.5}
		\def\cnnto{11.0}
		\def\cnntt{-0.5}
		\begin{tikzpicture}
		\tikzstyle{neuron}=[circle, draw=black, fill=white, minimum size=10pt, inner sep=0pt]
		
		\node[inner sep=0pt, label=north:\small{Input image $x$}] (frog) at (0,0)
		{\includegraphics[width=.1\textwidth]{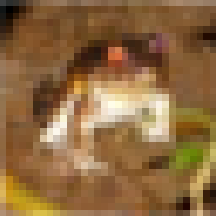}};
		
		\node (N1) at (2.1,0) {};
		\draw[->, thick] (frog) -- (N1);
		
		\node[draw, rectangle, fill=white, minimum width=1.5cm, minimum height=1.5cm, label=north:\small{Augmentation CNN}] (CNN1) at (2.75,0.25) {};
		%\draw[fill=white] (\cnnoo,\cnnot) rectangle ++(\cnnsize,\cnnsize);
		\draw[fill=white] (\cnnoo +0.2,\cnnot -0.2) rectangle ++(\cnnsize,\cnnsize);
		\draw[fill=white] (\cnnoo +0.4,\cnnot -0.4) rectangle ++(\cnnsize,\cnnsize);
		\draw[fill=white] (\cnnoo +0.6,\cnnot -0.6) rectangle ++(\cnnsize,\cnnsize);
		
		\node (N2) at (4.0,0) {};
		\node (N3) at (5.1,0) {};
		\draw[->, thick] (N2) -- (N3) node[midway, anchor=south] {\footnotesize{flatten}};
		
		% Draw the flattened layer nodes
		\node[neuron] (F1) at (5.2,0.8) {};
		\node[neuron] (F2) at (5.2,0.3) {};
		\node[neuron] (F3) at (5.2,-0.7) {};
		\path (F2) -- (F3) node [black, font=\small, pos=0.35] {$\vdots$};
		
		\node[draw, rectangle, rounded corners, minimum width=0.5cm, minimum height=1.3cm, fill=green!25, label={[label distance=-0.1cm]45:\small{$\bm{p}(x)$}}] (R1) at (6.6, 1.4) {};
		\node[draw, rectangle, rounded corners, minimum width=0.5cm, minimum height=1.3cm, fill=blue!25, label={[label distance=-0.1cm]0:\small{$\bm{m}(x)$}}] (R2) at (6.6, 0) {};
		\node[draw, rectangle, rounded corners, minimum width=0.5cm, minimum height=1.3cm, fill=orange!25, label={[label distance=-0.1cm]-45:\small{$\bm{\pi}(x)$}}] (R3) at (6.6, -1.4) {};
		
		% Draw the augment output layer nodes
		\node[neuron] (A1) at (6.6,1.8) {};
		\node[neuron] (A2) at (6.6,1.0) {};
		\path (A1) -- (A2) node [black, font=\small, pos=0.3] {$\vdots$};
		\node[neuron] (A3) at (6.6,0.4) {};
		\node[neuron] (A4) at (6.6,-0.4) {};
		\path (A3) -- (A4) node [black, font=\small, pos=0.3] {$\vdots$};
		\node[neuron] (A5) at (6.6,-1.0) {};
		\node[neuron] (A6) at (6.6,-1.8) {};
		\path (A5) -- (A6) node [black, font=\small, pos=0.3] {$\vdots$};
		
		\path (F1) edge (A1);
		\path (F1) edge (A2);
		\path (F1) edge (A3);
		\path (F1) edge (A4);
		\path (F1) edge (A5);
		\path (F1) edge (A6);
		\path (F2) edge (A1);
		\path (F2) edge (A2);
		\path (F2) edge (A3);
		\path (F2) edge (A4);
		\path (F2) edge (A5);
		\path (F2) edge (A6);
		\path (F3) edge (A1);
		\path (F3) edge (A2);
		\path (F3) edge (A3);
		\path (F3) edge (A4);
		\path (F3) edge (A5);
		\path (F3) edge (A6);
		
		\node[inner sep=0pt, label=south:\small{Original Image}] (frog2) at (9.0,-1.3)
		{\includegraphics[width=.1\textwidth]{frog.png}};
		
		\node[inner sep=0pt, label=north:\small{Augmented Image}] (frog_aug) at (9.0,1.3)
		{\includegraphics[width=.1\textwidth]{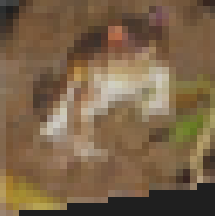}};
		
		\draw[->, thick] (frog2) -- (frog_aug);
		\draw[->, thick] (R1) -- (frog_aug);
		\draw[->, thick] (R2) -- (frog_aug);
		\draw[->, thick] (R3) -- (frog_aug);
		
		\node[draw, rectangle, fill=white, minimum width=1.5cm, minimum height=1.5cm, label=north:\small{Classification CNN}] (CNN2) at (11.75,0.25) {};
		%\draw[fill=white] (\cnnto,\cnntt) rectangle ++(\cnnsize,\cnnsize);
		\draw[fill=white] (\cnnto +0.2,\cnntt -0.2) rectangle ++(\cnnsize,\cnnsize);
		\draw[fill=white] (\cnnto +0.4,\cnntt -0.4) rectangle ++(\cnnsize,\cnnsize);
		\draw[fill=white] (\cnnto +0.6,\cnntt -0.6) rectangle ++(\cnnsize,\cnnsize);
		
		\draw[->, thick] (frog_aug.east) -- (CNN2.west);
		
		\node (N4) at (13.0,0) {};
		\node (N5) at (14.1,0) {};
		\draw[->, thick] (N4) -- (N5) node[midway, anchor=south] {\footnotesize{flatten}};
		
		% Draw the flattened layer nodes
		\node[neuron] (F21) at (14.2,0.8) {};
		\node[neuron] (F22) at (14.2,0.3) {};
		\node[neuron] (F23) at (14.2,-0.7) {};
		\path (F22) -- (F23) node [black, font=\small, pos=0.35] {$\vdots$};
		
		\node[draw, rectangle, rounded corners, minimum width=0.5cm, minimum height=2cm, fill=white, label=north:\small{Output}] (R4) at (15.6, 0.025) {};
		
		% Draw the output layer nodes
		\node[neuron] (O1) at (15.6,0.8) {};
		\node[neuron] (O2) at (15.6,0.3) {};
		\node[neuron] (O3) at (15.6,-0.7) {};
		\path (O2) -- (O3) node [black, font=\small, pos=0.35] {$\vdots$};
		
		\path (F21) edge (O1);
		\path (F21) edge (O2);
		\path (F21) edge (O3);
		\path (F22) edge (O1);
		\path (F22) edge (O2);
		\path (F22) edge (O3);
		\path (F23) edge (O1);
		\path (F23) edge (O2);
		\path (F23) edge (O3);
		\end{tikzpicture}}
	\caption{The overall architecture of the DNA model. During the policy search and classifier training phases, input images are fed through an augmentation network that outputs an input-dependent augmentation policy $(\bm{p}(x),\bm{m}(x),\bm{\pi}(x))$. Note that $\bm{p}(x)$ and $\bm{m}(x)$ are sigmoid outputs while $\bm{\pi}(x)$ is a softmax output to be consistent with the parameterization of the search space. The policy is then applied to the original image, and the resulting augmented version is fed through a classifier network to obtain the final output. Both networks are trained during the search phase, but during classifier training  the augmentation network remains frozen. During testing, input images are fed through the classifier network only.}
	\label{fig:DNA}
\end{figure*}

At a high level, DNA uses a policy search space whose augmentation parameters are functions of the model's input. We use a neural network to approximate these functions and then augment the data accordingly before feeding it through a separate network for classification. The augmentation network and classification network are jointly optimized as one end-to-end model during the policy search phase, before the weights of the augmentation network are frozen. This network then serves as an input-conditional DA policy used to train the classification network. The entire model architecture is shown in \Cref{fig:DNA}.

DNA shares some similarities with dynamic neural networks, which are networks whose weights are functions of the input \cite{HHSYWW21}. In both cases, the operation performed on the input varies from one instance to another, although it is only the augmentation that is input-dependent in our DNA model. Hence, our model allows for truly dynamic augmentation policies that can learn to apply different transformations depending on features of the input data.

%%%%%%%%% FIGURE %%%%%%%%%

\subsection{Policy Search Space}

We define our policy similar to AA \cite{CZMVL19} and other optimization-based automatic DA methods. Let $\mathcal{O}$ be a predetermined set of transformations $O: \mathcal{X} \to \mathcal{X}$, where $\mathcal{X}$ is some fixed image space. Each image operation $O$ has a parameter $m \in [0,1]$ which dictates the magnitude of the operation (e.g., for rotation, $m \in [0,1]$ corresponds to how many degrees to rotate the image by in the range $[-180^\circ,180^\circ]$). Some operations (e.g., horizontal flip) do not use the magnitude parameter but still have one. We introduce another parameter $p \in [0,1]$ and define a probabilistic operation $\bar{O}$ which corresponds to $O$ with probability $p$ and no operation with probability $1-p$. The difference in the augmentation policies used in DNA, compared to existing approaches, is that the augmentation parameters $p$ and $m$ are functions of the input $x$. Thus, a single operation in DNA is given by
\begin{equation}
    \bar{O}(x;p(x),m(x)) = \begin{cases}
        O(x;m(x))   &\text{with prob. } p(x) \\
        x           &\text{with prob. } 1-p(x).
    \end{cases}
\end{equation}
A sub-policy $s$ is defined as a composition of $k$ such operations
\begin{equation}
    x_{(\ell)} = \bar{O}^s_{\ell}(x_{(\ell-1)}; p^s_\ell(x), m^s_\ell(x)), \hspace{0.5cm} \ell=1,\dots,k
\end{equation}
where $x_{(0)} = x$ and $x_{(k)} = s(x;\bm{p}^s(x),\bm{m}^s(x))$. Notice that the parameters $p^s_\ell$ and $m^s_\ell$ only depend on the original input $x$ even though the operations $\bar{O}^s_\ell$ are applied sequentially.

We fix $\mathcal{O}$ to include the same image operations used in AA: ShearX/Y, TranslateX/Y, Rotate, AutoContrast, Invert, Equalize, Solarize, Posterize, Contrast, Color, Brightness, Sharpness, and Cutout (the only AA operation we exclude is Sample Pairing \cite{I18} because it uses two training images). As in AA, we use sub-policies of length $k=2$ operations. The entire list of sub-policies consists of all $\binom{15}{2}=105$ combinations of these 15 operations (each combination only appears once; e.g., (`TranslateX',`Rotate') is a sub-policy but (`Rotate',`TranslateX') is not).

Finally, a complete DNA augmentation policy includes an input-dependent probability vector $\bm{\pi}(x)$ over the space of all sub-policies. The value $\bm{\pi}_s(x)$ indicates the probability that sub-policy $s$ is applied to $x$ during the augmentation step while training the DNA model. An entire DA policy is defined uniquely by its parameter functions $(\bm{p}(x),\bm{m}(x),\bm{\pi}(x))$. The DNA model seeks to optimize this policy during the policy search phase by approximating $\bm{p}(x)$, $\bm{m}(x)$, and $\bm{\pi}(x)$ with a deep neural network as depicted in \Cref{fig:DNA}.

\subsection{Policy Sampling}

DNA augmentation requires sampling from a categorical distribution with parameter vector $\bm{\pi}(x)$, as well as sampling from $k$ Bernoulli distributions with parameters $p^s_1(x), \dots,p^s_k(x)$ to determine whether an operation should be applied. However, sampling from these distributions is not differentiable. Following DADA \cite{LHWHRY20}, we introduce differentiable relaxations of these sampling operations so that the entire DNA model is differentiable.

To make the sampling operations differentiable, we first write the selected sub-policy as
\begin{equation}
    s^*(x) = \sum_{s \in \mathcal{S}} c^s_x s(x),
\end{equation}
where $\mathcal{S}$ is the set of all sub-policies and $c_x$ is a one-hot vector sampled from the categorical distribution with probability vector $\bm{\pi}(x)$. In the DNA model, $\bm{\pi}(x)$ is given by a softmax output
\begin{equation} \label{eq:softmax}
    \bm{\pi}_s(x) = \frac{\exp(\alpha_s(x))}{\sum_{s' \in \mathcal{S}} \exp(\alpha_{s'}(x))}
\end{equation}
where $\alpha(x)$ is the output prior to the softmax layer. Sampling directly from this distribution would require backpropagating with respect to the parameters of the distribution. Instead, we can approximate $\bm{\pi}(x)$ with a Gumbel-Softmax \cite{JGP17} distribution
\begin{equation} \label{eq:gumbel}
    \bm{\pi}_s(x) \approx \frac{\exp((\alpha_s(x)+g_s)/\tau)}{\sum_{s' \in \mathcal{S}} \exp((\alpha_{s'}(x)+g_{s'})/\tau))}
\end{equation}
where $g_s = -\log(-\log(u_s))$ with $u_s \stackrel{i.i.d.}{\sim} \text{Uniform}(0,1)$, and $\tau$ is a temperature parameter. We now have the relaxation $c_x = \text{one-hot}(\argmax_s(\alpha_s(x)+g_s))$. This is an example of a reparameterization trick since $g_s$ has a fixed distribution unlike $\bm{\pi}_s(x)$. While the $\argmax$ operation is still not differentiable, we can use this sampling in the model's forward pass and backpropagate with respect to the deterministic parameters in \Cref{eq:gumbel}. Here we are using the straight-through gradient estimator \cite{BLC13} by effectively ignoring the $\argmax$ during backpropagation.

Once a sub-policy $s$ has been selected, we apply the sequence of operations $\bar{O}^s_1,\dots,\bar{O}^s_k$ in that sub-policy. We can write
\begin{equation} \label{eq:sample}
    \bar{O}^s_i(x;p^s_i(x),m^s_i(x)) = bO^s_i(x;m^s_i(x)) + (1-b)x
\end{equation}
where $b \sim \text{Bernoulli}(p^s_i(x))$. We again cannot sample directly from this distribution, but we use a reparameterization trick
\begin{equation} \label{eq:bernoulli}
    \hat{b} \approx \sigma \left( \lambda^{-1} \left( \log \frac{p^s_i(x)}{1-p^s_i(x)} + \log \frac{u^s_i}{1-u^s_i} \right) \right)
\end{equation}
where $\sigma$ is the sigmoid function, $\lambda$ is a temperature parameter, and $u^s_i \stackrel{i.i.d.}{\sim} \text{Uniform}(0,1)$. We set $b = \mathbbm{1}\{\hat{b}>0.5\}$ during the forward pass but use the straight-through gradient estimator to backpropagate with respect to the parameters in \Cref{eq:bernoulli}. Finally, we use the straight-through gradient estimator for $m^s_i$ in \Cref{eq:sample} since some of the image operations $O^s_i$ (e.g., horizontal flipping) are not differentiable.

\section{Experiments and Results}

We assess our DNA model on the CIFAR-10 \cite{K09}, CIFAR-100 \cite{K09}, and SVHN \cite{NWCBWN11} % and Imagenet [cite]
datasets and compare the results to other automated data augmentations models (namely AA \cite{CZMVL19}, PBA \cite{HLSAC19}, Fast AA \cite{LKKK19}, and DADA \cite{LHWHRY20}). Using DNA involves a search phase and a training phase. During the search phase, we jointly optimize the augmentation network and classification network so that the model can learn an effective input-conditional augmentation strategy. After the policy search is complete, the weights in the augmentation network are frozen. During training, the augmentation network is used to augment each input image, but this network is not updated. Instead, the classification network receives the augmented inputs and is trained just like any classifier would normally be on an augmented dataset. Finally, we evaluate the trained model on test data and use test accuracy to determine whether the augmentation network learned an effective DA policy that improves generalization. Our implementation uses some existing code assets from Fast AA and DADA for the policy search space and augmentation but primarily uses original code for model specification and training.

%%%% TABLE %%%%
\begin{table}
    \centering
    %\resizebox{\linewidth}{!}{
    \begin{tabular}{lccccc} \hline
        Dataset & AA \cite{CZMVL19} & PBA \cite{HLSAC19} & Fast AA \cite{LKKK19} & DADA \cite{LHWHRY20} & DNA \\ \hline
        CIFAR-10 & 5000 & 5 & 3.5 & 0.1 & 0.4 \\
        CIFAR-100 & -- & -- & -- & 0.2 & 0.5 \\
        SVHN & 1000 & 1 & 1.5 & 0.1 & 0.1 \\ \hline
    \end{tabular}%}
    \caption{GPU hours spent on DA policy search. Hours for AA, PBA, Fast AA, and DADA are reported in their respective papers. Hours for DNA show the search cost using an NVIDIA V100 GPU.}
    \label{tab:gpu_hours}
\end{table}

\paragraph{Search Phase}

We conduct DA policy search by optimizing the augmentation network on reduced versions of each dataset. Following AA, we randomly select 4,000 training examples for the reduced CIFAR-10 and CIFAR-100 datasets and 1,000 samples for the reduced SVHN dataset. In each case, we use stratified sampling to preserve the percentage of each class in the dataset. CIFAR data are also pre-processed using horizontal flips with 50\% probability, zero-padding with random crops, and Cutout \cite{DT17} with size $16 \times 16$ patches, but on SVHN we only apply Cutout. This pre-processing follows AA and provides a baseline for effective training on these datasets. We conduct policy search for 20 epochs on all datasets.

During the search phase, we use a Wide-ResNet-40-2 \cite{ZK16} architecture for CIFAR data and a Wide-ResNet-28-10 architecture for SVHN (with the same architecture for both the augmentation network and the classification network). The augmentation network is trained using an Adam optimizer \cite{KB17} with learning rate $0.005$, momentum $\beta=(0.5,0.999)$, and weight decay of $0$. The classification network is trained using an SGD optimizer with momentum $0.9$, weight decay $2 \times 10^{-4}$, and a cosine annealing learning rate schedule. The initial learning rate and batch size are set to $0.1$ and $128$ for CIFAR-10, compared to $0.025$ and $32$ for CIFAR-100 and SVHN. Finally, temperature parameters for the relaxed categorical and Bernoulli distributions were set to $\tau=\lambda=0.5$.

\Cref{tab:gpu_hours} shows the cost of DA policy search in GPU hours for each model on each dataset. Note that AA, PBA, and Fast AA did not perform policy search on CIFAR-100 due to the computationally intensive nature of their algorithms. They instead transferred the DA policy found during CIFAR-10 search to train their models on CIFAR-100. 

Most notably, DNA requires much less search time than AA, PBA, and Fast AA since those methods use expensive techniques like reinforcement learning and population based optimization. Policy search is faster with DNA by orders of magnitude because it benefits from the single-pass differentiable nature of the model as with DADA. DNA requires similar or slightly more time for policy search than DADA, likely because the input-conditional augmentation policy is learned with a neural network rather than a single set of parameters.

%%%% TABLE %%%%
\begin{table*}
	\centering
	%\resizebox{\linewidth}{!}{
	\begin{tabular}{l|ccccc|c} \hline
		Dataset & Baseline & Cutout \cite{DT17} & AA \cite{CZMVL19} & Fast AA \cite{LKKK19} & DADA \cite{LHWHRY20} & DNA \\ \hline
		CIFAR-10 & 5.3 & 4.1 & 3.7 & 3.6 & 3.6 & 4.0 \\
		CIFAR-100 & 26.0 & 25.2 & 20.7 & 20.7 & 20.9 & 22.0 \\
		SVHN & 1.5 & 1.3 & 1.1 & 1.1 & 1.2 & 3.4 \\ \hline
	\end{tabular}%}
	\caption{Test error rates (\%) on each dataset using a Wide-ResNet-40-2 network architecture for CIFAR data and a Wide-ResNet-28-10 for SVHN. Baseline refers to conventional pre-processing augmentation without Cutout. Error rates for AA, Fast AA, and DADA are reported in their respective papers. Error rates for DNA are computed on the standard CIFAR-10, CIFAR-100, and SVHN test datasets. PBA \cite{HLSAC19} does not use the same architecture and hence is not compared.}
	\label{tab:error}
\end{table*}

\paragraph{Training Phase} After DA policy search is conducted, the weights of the augmentation network are frozen. This gives us a learned input-conditional augmentation policy that can be used to augment training data. During training, each image is still fed through the augmentation network but without updating the network's weights. The resulting policy is used to augment the original image, which is fed through the classification network for the final output. We train our DNA models on the full CIFAR-10, CIFAR-100, and SVHN training sets (the ``core'' SVHN training set, not the ``extra'' one) for 200 epochs. Again, all training data is pre-processed as in the search phase to be consistent with AA and other DA literature on the datasets.

AA, Fast AA, and DADA sometimes refer to this training phase as policy evaluation (and distinguish it from model evaluation that occurs at test time). This idea of policy evaluation is especially relevant for AA, PBA, and Fast AA since performance on training data is used as a signal for the overall optimization scheme. On the other hand, DADA and DNA use a one-pass optimization strategy since the model is fully differentiable.   

During training, both networks again use a Wide-ResNet-40-2 architecture for CIFAR and a Wide-ResNet-28-10 architecture for SVHN. We use an SGD optimizer with the same parameters as the search phase, except the initial learning rate is adjusted to $0.1$ for CIFAR and $0.01$ for SVHN and the batch size is adjusted to $512$ for all datasets, and we add Nesterov momentum \cite{BLB17}. This optimizer is applied only to the classification network. All training parameters and the number of epochs trained are consistent with those used in AA, Fast AA, and DADA. 

\paragraph{Model Testing} After DA policy search and training have been completed, we test our DNA model on the standard test datasets for CIFAR-10, CIFAR-100, and SVHN. The test images are not augmented in any way, nor are they fed through the augmentation network. Augmentation was used during training with the goal of exposing the model to new data and improving generalization. At test time, images are fed straight through the classification network.

\Cref{tab:error} shows error rates on each test dataset for baseline pre-processing, Cutout, AA, Fast AA, DADA, and DNA using the same network architecture. Each of the automated data augmentation algorithms AA, Fast AA, and DADA perform better than baseline pre-processing and Cutout, which are known to perform very well on these natural image datasets. Our DNA model performs better than baseline pre-processing and Cutout but not quite as well as the other DA methods. We believe that optimization of our training hyperparameters could improve our error rates to be more competitive with the other methods, as the different structure of our augmentation policy likely warrants different hyperparameters.

\subsection{Input-Conditional Augmentation Policies}

\Cref{tab:gpu_hours,tab:error} show that our DNA model conducts DA policy search very efficiently and generalizes well on test data. One additional benefit of DNA is that it learns an input-conditional DA policy. By investigating aggregate statistics of the found augmentation policies over the entire training set, we can gain insight into which operations are beneficial for certain images or classes.

Part of the motivation for DNA's input-conditional augmentation is the idea that different class labels may be invariant to different transformations. To investigate whether this phenomenon occurs in natural image data, we look at the average DA policy for each class in the CIFAR-10 training set. In particular, for a given class label $y$, we look at the average probability vector $\mathbb{E}[\bm{\pi}(x) | Y=y]$. Of interest are the five largest values in this vector, corresponding to the five most commonly used sub-policies for class $y$. \Cref{tab:policies} shows the top five sub-policies averaged over each class in the CIFAR-10 training set compared to the sub-policies found by AA on CIFAR-10.

%%%% TABLE %%%%
\begin{table*}
	\centering
	\resizebox{\linewidth}{!}{
		\begin{tabular}{l|l|l|l|l|l|l} \hline
			& airplane & automobile & bird & cat & deer & AA \cite{CZMVL19} \\ \hline
			sub-policy 1 & (TranslateX, Solarize) & (TranslateX, Solarize) & (TranslateX, Solarize) & (TranslateX, Solarize) & (TranslateX, Solarize) & (Invert, Contrast) \\
			sub-policy 2 & (Equalize, Posterize) & (Equalize, Posterize) & (Equalize, Posterize) & (Equalize, Posterize) & (Equalize, Posterize) & (Rotate, TranslateX) \\
			sub-policy 3 & (ShearY, Contrast) & (ShearY, Contrast) & (TranslateY, Contrast) & (TranslateY, Contrast) & (TranslateY, Contrast) & (Sharpness, Sharpness) \\
			sub-policy 4 & (Equalize, Brightness) & (Equalize, Brightness) & (Invert, Cutout) & (Invert, Cutout) & (Invert, Cutout) & (ShearY, TranslateY) \\
			sub-policy 5 & (Invert, Cutout) & (Invert, Cutout) & (ShearY, Contrast) & (ShearY, Contrast) & (ShearY, Contrast) & (AutoContrast, Equalize) \\ \cline{1-6}
			& dog & frog & horse & ship & truck & (ShearY, Posterize) \\ \cline{1-6} 
			sub-policy 1 & (TranslateX, Solarize) & (TranslateX, Solarize) & (TranslateX, Solarize) & (TranslateX, Solarize) & (TranslateX, Solarize) & (Color, Brightness) \\
			sub-policy 2 & (Equalize, Posterize) & (Equalize, Posterize) & (Equalize, Posterize) & (ShearY, Contrast) & (Equalize, Posterize) & (Sharpness, Brightness) \\
			sub-policy 3 & (TranslateY, Contrast) & (TranslateY, Contrast) & (TranslateY, Contrast) & (Equalize, Posterize) & (ShearY, Contrast) & (Equalize, Equalize) \\
			sub-policy 4 & (Invert, Cutout) & (Invert, Cutout) & (Invert, Cutout) & (Equalize, Brightness) & (Equalize, Brightness) & (Contrast, Sharpness) \\
			sub-policy 5 & (ShearY, Contrast) & (TranslateX, Posterize) & (ShearY, Contrast) & (Invert, Cutout) & (Invert, Cutout) & (Color, TranslateX) \\ \hline 
	\end{tabular}}
	\caption{Ordered top five DNA sub-policies for each CIFAR-10 class compared to ordered top eleven sub-policies for AA on all of CIFAR-10. Sub-policies are ordered by the probability $\pi$ of sampling that sub-policy from largest to smallest, averaged over each class for DNA.}
	\label{tab:policies}
\end{table*}

\Cref{tab:policies} offers some interesting insights about the input-conditional augmentation policies learned by our DNA model on CIFAR-10. The sub-policy (TranslateX, Solarize) is most common for each class, followed by (Equalize, Posterize) for every class except ship. After the top two sub-policies there are some additional differences across classes, but in total we only see seven unique sub-policies appear in the top five across all CIFAR-10 classes. Thus, it appears that DNA learned a few sub-policies that improve generalization for all CIFAR-10 classes while also picking up some minor differences in DA policies between classes.

Notably, DNA appears to have learned two groups of DA policies -- one on images of animals and one on images of vehicles. The top five sub-policies are identical for the bird, cat, deer, dog, and horse classes. Additionally, only sub-policy 5 is different between these classes and the frog class, suggesting that images of animals benefit from similar augmentations. On the other hand, the top five sub-policies are identical for the airplane, automobile, and truck classes. Moreover, the ship class only differs in the order of sub-policies 2 and 3. Overall, the animal classes have very similar sub-policies as do the vehicle classes, and there are a few noticeable differences between these two groups in sub-policies 3-5.

Our DNA model favors color-based transformations over geometric transformations among the top five sub-policies on CIFAR-10. Color-based augmentation is often preferred for natural image data, and this agrees with AA, which also finds fewer geometric transformations among its top sub-policies. In total, a breakdown of the DNA augmentation policies by class gives a more thorough picture of which augmentations are beneficial for certain classes or images.

\section{Discussion}

DNA offers a more general model for automated data augmentation that allows for input-conditional augmentation policies. While this approach is compelling and potentially very robust to diverse datasets, it could have some negative societal impacts. Most notably, if the DNA model is deployed on data with historical bias, it is conceivable that it could learn those biases even more than other models due to the input-dependent nature of its data augmentation. To mitigate this concern, we encourage continued awareness in the machine learning community about biases present in data and processes to avoid misuse.

Our results show that DNA does not perform quite as well as other automated DA approaches, suggesting some limitations of the model. It is quite possible that the input-conditional augmentation policy's additional complexity makes the learning process slightly slower. Moreover, on datasets when augmentation would not benefit from being input-dependent, it is likely that DNA provides little additional benefit over DADA. However, we believe that our model gives a compelling foundation for automated conditional data augmentation that could improve model performance and generalization, especially on more varied datasets. We plan to further experiment with DNA using different training hyperparameters and datasets to better understand the effect that conditional data augmentation has on downstream model performance.

\section{Conclusions}

In this paper we propose Dynamic Network Augmentation (DNA), a method for learning input-conditional data augmentation policies from training data alone. The DNA architecture features a differentiable end-to-end model where one network is trained to learn augmentation policy functions and another network is trained as a classifier. 

While DNA does not outperform comparable models in terms of accuracy, its novel ability to efficiently search for input-conditional policies provides new opportunities for data-driven data augmentation strategies. Investigating the top sub-policies learned from CIFAR-10 by class reveals that DNA is choosing more color-based transformations, which is consistent with other augmentation strategies on natural images. Grouping instances by sub-policy also reveals that our learned augmentations correlate with natural semantic boundaries (e.g., animal vs.\ vehicle), suggesting that DNA learns meaningful and insightful DA policies that fit a particular dataset.

\bibliographystyle{plain}
\bibliography{references}

\begin{thebibliography}{10}

\bibitem{BLC13}
Yoshua Bengio, Nicholas L\'{e}onard, and Aaron Courville.
\newblock Estimating or propagating gradients through stochastic neurons for
  conditional computation.
\newblock arXiv:1308.3432v1, 2013.

\bibitem{BFIW20}
Gregory Benton, Marc Finzi, Pavel Izmailov, and Andrew~Gordon Wilson.
\newblock Learning invariances in neural networks.
\newblock In {\em Advances in Neural Information Processing Systems (NeurIPS)},
  pages 1--16, 2020.

\bibitem{BLB17}
Aleksandar Botev, Guy Lever, and David Barber.
\newblock Nesterov's accelerated gradient and momentum as approximations to
  regularised update descent.
\newblock In {\em International Joint Conference on Neural Networks (IJCNN)},
  2017.

\bibitem{CW16a}
Taco~S. Cohen and Max Welling.
\newblock Group equivariant convolutional networks.
\newblock In {\em Proceedings of the 33rd International Conference on Machine
  Learning (ICML)}, volume~48, pages 2990--2999, 2016.

\bibitem{CW16b}
Taco~S. Cohen and Max Welling.
\newblock Steerable {CNNs}.
\newblock arXiv:1612.08498, 2016.

\bibitem{CZMVL19}
Ekin~D. Cubuk, Barret Zoph, Dandelion Man\'{e}, Vijay Vasudevan, and Quoc~V.
  Le.
\newblock Autoaugment: Learning augmentation strategies from data.
\newblock In {\em Proceedings of IEEE Conference on Computer Vision and Pattern
  Recognition (CVPR)}, pages 113--123, 2019.

\bibitem{DT17}
Terrance DeVries and Graham~W. Taylor.
\newblock Improved regularization of convolutional neural networks with cutout.
\newblock arXiv:1708.04552v2, 2017.

\bibitem{GPMXWOCB14}
Ian~J. Goodfellow, Jean Pouget-Abadie, Mehdi Mirza, Bing Xu, David
  Warde-Farley, Sherjil Ozair, Aaron Courville, and Yoshua Bengio.
\newblock Generative adversarial nets.
\newblock In {\em Advances in Neural Information Processing Systems (NeurIPS)},
  pages 2672--2680, 2014.

\bibitem{HHSYWW21}
Yizeng Han, Gao Huang, Shiji Song, Le~Yang, Honghui Wang, and Yulin Wang.
\newblock Dynamic neural networks: A survey.
\newblock arXiv:2102.04906v3, 2021.

\bibitem{HLSAC19}
Daniel Ho, Eric Liang, Ion Stoica, Pieter Abbeel, and Xi~Chen.
\newblock Population based augmentation: Efficient learning of augmentation
  policy schedules.
\newblock In {\em Proceedings of the 36th International Conference on Machine
  Learning (ICML)}, pages 2731--2741, 2019.

\bibitem{I18}
Hiroshi Inoue.
\newblock Data augmentation by pairing samples for images classification.
\newblock arXiv:1801.02929v2, 2018.

\bibitem{JGP17}
Eric Jang, Shixiang Gu, and Ben Poole.
\newblock Categorical reparameterization with gumbel-softmax.
\newblock In {\em International Conference on Learning Representations (ICLR)},
  2017.

\bibitem{KB17}
Diederik~P. Kingma and Jimmy~L. Ba.
\newblock Adam: A method for stochastic optimization.
\newblock arXiv:1412.6980v9, 2017.

\bibitem{K09}
Alex Krizhevsky.
\newblock Learning multiple layers of features from tiny images.
\newblock Technical report, University of Toronto, 2009.

\bibitem{KSH12}
Alex Krizhevsky, Ilya Sutskever, and Geoffrey~E. Hinton.
\newblock Imagenet classification with deep convolutional neural networks.
\newblock In {\em Advances in Neural Information Processing Systems (NeurIPS)},
  pages 1097--1105, 2012.

\bibitem{LB98}
Yann LeCun and Yoshua Bengio.
\newblock {\em Convolutional Networks for Images, Speech, and Time Series},
  pages 255--258.
\newblock MIT Press, Cambridge, MA, USA, 1998.

\bibitem{LCB10}
Yann LeCun, Corinna Cortes, and CJ~Burges.
\newblock {MNIST} handwritten digit database.
\newblock {\em ATT Labs [Online]. Available: http://yann.lecun.com/exdb/mnist},
  2, 2010.

\bibitem{LBC17}
Joseph Lemley, Shabab Bazrafkan, and Peter Corcoran.
\newblock Smart augmentation learning an optimal data augmentation strategy.
\newblock {\em IEEE Access}, 5:5858--5869, 2017.

\bibitem{LHWHRY20}
Yonggang Li, Guosheng Hu, Yongtao Wang, Timothy Hospedales, Neil~M. Robertson,
  and Yongxin Yang.
\newblock Differentiable automatic data augmentation.
\newblock In {\em Proceedings of the European Conference on Computer Vision
  (ECCV)}, pages 580--595, 2020.

\bibitem{LKKK19}
Sungbin Lim, Ildoo Kim, Taesup Kim, and Chiheon Kim.
\newblock Fast autoaugment.
\newblock In {\em Advances in Neural Information Processing Systems (NeurIPS)},
  pages 6665--6675, 2019.

\bibitem{LSY19}
Hanxiao Liu, Karen Simonyan, and Yiming Yang.
\newblock Darts: Differentiable architecture search.
\newblock In {\em International Conference on Learning Representations (ICLR)},
  2019.

\bibitem{NWCBWN11}
Yuval Netzer, Tao Wang, Adam Coates, Alessandro Bissacco, Bo~Wu, and Andrew~Y.
  Ng.
\newblock Reading digits in natural images with unsupervised feature learning.
\newblock In {\em NIPS Workshop on Deep Learning and Unsupervised Feature
  Learning}, 2011.

\bibitem{SPTSWW17}
Ashish Shrivastava, Tomas Pfister, Oncel Tuzel, Josh Susskind, Wenda Wang, and
  Russ Webb.
\newblock Learning from simulated and unsupervised images through adversarial
  training.
\newblock In {\em Proceedings of IEEE Conference on Computer Vision and Pattern
  Recognition (CVPR)}, pages 2107--2116, 2017.

\bibitem{SSP03}
P.Y. Simard, D.~Steinkraus, and J.C. Platt.
\newblock Best practices for convolutional neural networks applied to visual
  document analysis.
\newblock In {\em Proceedings of the 7th International Conference on Document
  Analysis and Recognition (ICDAR)}, pages 958--963, 2003.

\bibitem{SZ15}
Karen Simonyan and Andrew Zisserman.
\newblock Very deep convolutional networks for large-scale image recognition.
\newblock In {\em Advances in Neural Information Processing Systems (NeurIPS)},
  2015.

\bibitem{SLJSRAEVR15}
Christian Szegedy, Wei Liu, Yangqing Jia, Pierre Sermanet, Scott Reed, Dragomir
  Anguelov, Dumitru Erhan, Vincent Vanhoucke, and Andrew Rabinovich.
\newblock Going deeper with convolutions.
\newblock In {\em Proceedings of IEEE Conference on Computer Vision and Pattern
  Recognition (CVPR)}, pages 1--9, 2015.

\bibitem{TPCPR17}
Toan Tran, Trung Pham, Gustavo Carneiro, Lyle Palmer, and Ian Reid.
\newblock A bayesian data augmentation approach for learning deep models.
\newblock In {\em Advances in Neural Information Processing Systems (NeurIPS)},
  pages 2797--2806, 2017.

\bibitem{WC19}
Maurice Weiler and Gabriele Cesa.
\newblock General {E(2)}-equivariant steerable {CNN}s.
\newblock In {\em Advances in Neural Information Processing Systems (NeurIPS)},
  pages 14334--14345, 2019.

\bibitem{YHOCCY19}
Sangdoo Yun, Dongyoon Han, Seong~Joon Oh, Sanghyuk Chun, Junsuk Choe, and
  Youngjoon Yoo.
\newblock {CutMix:} regularization strategy to train strong classifiers with
  localizable features.
\newblock In {\em Proceedings of the IEEE/CVF International Conference on
  Computer Vision (ICCV)}, pages 6023--6032, 2019.

\bibitem{ZK16}
Sergey Zagoruyko and Nikos Komodakis.
\newblock Wide residual networks.
\newblock In {\em Proceedings of the British Machine Vision Conference (BMVC)},
  2016.

\bibitem{ZCDL18}
Hongyi Zhang, Moustapha Cisse, Yann~N. Dauphin, and David Lopez-Paz.
\newblock {mixup:} beyond empirical risk minimization.
\newblock arXiv:1710.09412v2, 2018.

\end{thebibliography}

\end{document}